\documentclass[a4paper]{article} 

\usepackage{blindtext} 
\usepackage{charter} 
\usepackage[utf8]{inputenc} 
\usepackage{microtype} 
\usepackage{amsthm, amsmath, amssymb} 
\usepackage{float} 
\usepackage[final, colorlinks = true, 
            linkcolor = black, 
            citecolor = black]{hyperref} 
\usepackage{graphicx, multicol} 
\usepackage{xcolor} 
\usepackage{marvosym, wasysym} 
\usepackage{rotating} 
\usepackage{censor} 
\usepackage{listings, style/lstlisting} 
\usepackage{pseudocode} 
\usepackage{style/avm} 
\usepackage{booktabs} 
\usepackage{tikz-qtree} 
\tikzset{every tree node/.style={align=center,anchor=north},
         level distance=2cm} 
\usepackage{style/btree} 
\usepackage{csquotes} 
\usepackage[yyyymmdd]{datetime} 
\usepackage{fancyhdr} 
\usepackage{multirow}
\usepackage{xurl}
\begin{document}


\fancyhead[C]{}
\hrule \medskip 
\begin{centering}
\large 
Towards Fairness Certification in Artificial Intelligence\\ 
\end{centering}
\normalsize 
\medskip\hrule 
\smallskip
\begin{center}
\begin{tabular}{ccc}
     \hspace{-1.5cm} \multirow{3}{*}{\includegraphics[width=4.5cm]{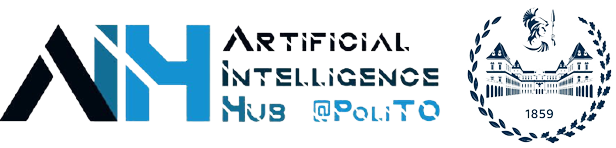}} &  
    Tatiana Tommasi$^1$, Silvia Bucci$^1$, Barbara Caputo$^1$, Pietro Asinari$^{1,2}$ & \multirow{3}{*}{\includegraphics[width=3.5cm]{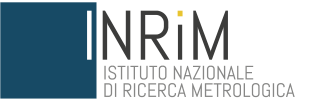}} \\
    &{$^1$ Politecnico di Torino}&\\
    &{$^2$ Istituto Nazionale di Ricerca Metrologica}&\\
\end{tabular}
\end{center}
\bigskip

Thanks to the great progress of machine learning in the last years, several Artificial Intelligence (AI) techniques have been increasingly moving from the controlled research laboratory settings to our everyday life. The most simple examples are the spam filters that keep our email account in order, face detectors that help us when taking a portrait picture, online recommender systems that suggest which movie and clothing we might like, or interactive maps that navigate us towards our vacation home. Artificial intelligence is clearly supportive in many decision-making scenarios, but when it comes to sensitive areas such as health care, hiring policies, education, banking or justice, with major impact on individuals and society, it becomes crucial to establish guidelines on how to design, develop, deploy and monitor this technology. Indeed the decision rules elaborated by machine learning models are data-driven and there are multiple ways in which discriminatory biases can seep into data. Algorithms trained on those data incur the risk of amplifying prejudices and societal stereotypes by over associating protected attributes such as gender, ethnicity or disabilities with the prediction task. 

\vspace{2mm}
The discussion about ethical principles for trustworthy AI has attracted the attention of the European Community which has recently published a legislative proposal for regulation of AI applications \cite{EUreg}. The goal is to guarantee that the most recent technologies are used in a way that is safe and compliant with the law, including the respect of fundamental rights. All the sensitive areas mentioned above appear in the proposal and the related AI systems are classified as high-risk with an explicit call for data quality and algorithmic robustness control against bias. In this scenario, standardisation would play a key role to define technical solutions that can be used by AI providers to ensure compliance to EU regulations.  

\vspace{2mm}
Starting from the extensive experience of the National Metrology Institute on measurement standards and certification roadmaps, and of Politecnico di Torino on machine learning as well as methods for domain bias evaluation and mastering, we propose a first joint effort to define the operational steps needed for AI fairness certification. Specifically we will overview the criteria that should be met by an AI system before coming into official service and the conformity assessment procedures useful to monitor its functioning for fair decisions.

\paragraph{State of the Art.}
Defining fairness is very challenging since it involves notions of social science, ethics and law. \emph{Machine learning fairness} has taken a simplified path by looking at the problem from the computational perspective. The aim is to ensure that data and models do not encode or enforce prejudice or favoritism towards an individual or a group based on their inherent or acquired characteristics. This task is mainly formalized by listing a set of protected variables such as gender, age or ethnic origin and evaluating first of all whether data are unbalanced with respect to them. An algorithm is said to be fair with respect to these attributes when its outcome does not rely on the information they convey. Of course the naive strategy of ignoring these attributes (aka \emph{fairness through unawareness}) does not lead to reliable solutions: machine learning and in particular deep learning models are able to discover implicit association of the protected variables with closely correlated features and rely on them for the final decision.  

\vspace{2mm}
Several statistical metrics have been defined to evaluate fairness \cite{du2020fairness,Oneto2020}. 
Considering a simple binary objective, we can indicate with $Y$ the label, with $\hat{Y}$ the model output and with $A$ the protected attribute in the data. The \emph{Demographic Parity} measure specifies that the positive prediction should be the same regardless of whether the sample is or not in the protected group ($P(\hat{Y}=1|A=0)=P(\hat{Y}=1|A=1)$). The \emph{Equality of Opportunity} checks whether the probability of a sample in a positive class being assigned to a positive outcome is equal for both protected and unprotected group members ($P(\hat{Y}=1|A=0,Y=1)=P(\hat{Y}=1|A=1,Y=1)$). The \emph{Equality of Odds} extends the previous metric to cover both the positive and negative class($Y=y, \quad y\in\{0,1\}$). The \emph{Predictive Parity} or \emph{Calibration} corresponds to the equality of opportunity but holds in the case of continuous model output $S$ by setting a threshold $\tau$  ($P(\hat{Y}=1|A=0,S>\tau)=P(\hat{Y}=1|A=1,S>\tau)$). In  more complex scenarios than binary prediction and also in case of multiple protected attributes, how to assess fairness becomes more complex and the evaluation largely benefit from research on \emph{Optimal Transport} \cite{pmlr-gordaliza19a,Silvia_Ray_Tom_Aldo_Heinrich_John_2020} and \emph{Causal Reasoning} \cite{pmlr-nabi19a,Oneto2020}.  

\vspace{2mm}
The current strategies for bias mitigation can be organized in three main groups \cite{mehrabi2019survey}. The \emph{pre-processing} techniques try to transform the data so that the underlying discrimination is removed \cite{aifairness360}. The \emph{in-processing} strategies modify state-of-the-art learning algorithms in order to remove discrimination during the model training process \cite{Adeli_2021_WACV,Wang_2020_CVPR}. This is generally obtained by including auxiliary objective functions to minimize the existing bias on the basis of one of the metrics described above. \emph{Post-processing} \cite{Oneto2020} is finally helpful in all those cases in which the learning algorithm cannot be modified but its output can be re-scored during deployment.

\paragraph{Current Gaps and Recommended Actions.}
From a preliminary analysis of the most recent machine learning fairness research results we draw few observations on existing procedural gaps and propose possible beneficial actions towards certification rules for AI system.

\begin{description}
  \item[\hspace{9mm}\textbf{Data.}] Although the EU proposal for regulation indicates the need for data governance, the metrics about quantity and suitability of the used data set are not specified. Depending on the considered task and related bias risk it would be helpful to have \begin{enumerate}
      \item a list of data aspects that must be evaluated. The protected attributes in each data collection should be accurately defined, paying attention to their categorical or continuous nature;
      \item a list of statistical hypothesis tests on each attribute and related confidence interval ranges considered safe for certification;
      \item a list of suggested methods to repair training data in order to enforce the attribute conditional independence that guarantees fairness (e.g. \cite{chuang2021fair,repairDB}). Quantifying the effect of those methods provides an indication of the effort needed to move the data towards the acceptable fairness level. 
  \end{enumerate}

  \item[\hspace{9mm}\textbf{Algorithms.}] Starting from the existing fairness metrics it is necessary to 
  \begin{enumerate}
  \item specify the extent to which the equations can be violated. Taking for instance the demographic parity, a bound should be declared on the conditional probability difference  $P(\hat{Y}=1|A=0)-P(\hat{Y}=1|A=1)<\epsilon$;
  \item sort by importance or exclude some of the criteria. Indeed it might be infeasible to satisfy all the criteria at once except in highly constrained cases \cite{kleinberg2017inherent}.
  \item assess the trade-off between accuracy and fairness. The maximum limit to which fairness conditions can compromise the model prediction accuracy should be declared, especially taking into consideration the accuracy for non-protected groups;
  \item maintain transparency. Interpretability and fairness should not be conflicting requirements and in particular it is crucial to set transparency rules that explain how fairness is met \cite{Quadrianto_2019_CVPR}.
  \end{enumerate}

  \item[\hspace{9mm}\textbf{Monitoring.}] The algorithms should be analyzed both in its training and deployment phase considering that
  \begin{enumerate}
  \item most of the learning procedures start from pre-trained models. As well as the data, the pre-trained models should pass a fairness screening that indicates how much they affect the final algorithm performance in the downstream task; 
  \item the impact of on individuals or groups may change over time as an effect of the adopted fairness safety measures \cite{pmlr-liu18}.  Thus it is important to maintain a temporal record and periodic update of the algorithm behavior;
  \item establishing the consequence of failure in novel unexpected situation is crucial. Depending on the intended target scenario, several stress-tests should be designed to assess the algorithm robustness (in terms of accuracy and fairness) under different level of domain shift \cite{DAsurvey}.
  \end{enumerate}
\end{description}

\bibliographystyle{plain}
\bibliography{reference}

\end{document}